# Automatic 3D object detection of Proteins in Fluorescent labeled microscope images with spatial statistical analysis

Ramin Norousi[a], Volker J. Schmid[a]

[a] Department of Statistics, Ludwig-Maximilians-University München, Germany

## ABSTRACT

Since manual object detection is very inaccurate and time consuming, some automatic object detection tools have been developed in recent years. At the moment, there is no image analysis software available which provides an automatic, objective assessment of 3D foci which is generally applicable. Complications arise from discrete foci which are very close or even come in contact to other foci, moreover they are of variable sizes and show variable signal-to-noise, and must be analyzed fully in 3D. Therefore we introduce the 3D-OSCOS (3D-Object Segmentation and Colocalization Analysis based on Spatial statistics) algorithm which is implemented as a user-friendly toolbox for interactive detection of 3D objects and visualization of labeled images.

## 1   INTRODUCTION

It is now widely understood that subcellular objects interact either in direct or indirect ways together to fulfill cellular functions. The interaction therefore depends on the spatial proximity which is essential for the direct molecular interaction or it depends on the distance between the structure for indirect interactions [Helmuth et al., 2010].

The detection of spots (subcellular objects) and their quantification in multichannel 3D Fluorescent images are generally accepted as key steps in order to understand how the spatial organization is established. Further they help us to gain local information of proteins distributed in nucleus. This information is crucial to describe the role of spot locations in the biological processes.

The manual Object detection task is very inaccurate and time consuming. Therefore an automatic analysis toolbox with the option of manual intervention in order to get a more accurate and flexible analysis is of great importance. This is the main objective of the project and the realized results show that the chosen and proved approach have been reasonable and professionally applied to deal with the mentioned requirements.

To prevent the laborious manual quantification, various automatic 2D and 3D approaches have been developed. First generation of automatic detection and quantification of subcellular objects was based on 2D approaches (slice-by-slice). Summarizing these methods, it can be said that the object segmentation and quantification step is performed among others by threshold-based [Fay, 1997], edge detection-based [Jaskolski et al., 2005], watershed transformation [Woolford, 2007], 2D Gaussian fit- ting [Trabesinger et al., 2001], which are performed in 2D (slice-based) without taking into consideration the expansion of the objects in 3D image.

Analyzing of colocalization between different 3D image channels based on 2D slice-based technique leads to inaccuracies and underestimates the number of colocalzations, since two colocalized objects could be visible in 2 adjacent slices but they could not be detected in one slice. Therefore a reasonable detection and quantification method should carry out the full information and reflect the nature of biological samples (e.g. proteins) in three dimensional spaces due to the fact that they are spanned across multiple slides and their three-dimensional nature and too. Moreover, due to the discrepancy between lateral and axial resolution of the confocal microscopy (e.g. 4 µm in x-, y- and 12 µm in z-direction), spots are usually distorted along the z-axis. Thus, for an accurate analysis, the degree of distortion by the optical device should be considered. Otherwise, the analyzing colocalization events in two-dimensional images leads to misinterpretation and incomplete spatial description state due to missing information about the size in z-direction. Facing these problems, several (semi-)automatic quantification and colocalization method images have been developed, whereas most of them are very problem specific. These great efforts were made to analyze spots by exploiting the full information of 3D structure data. Among others the following commonly used methods are described in more detail:

Worz et al. [Worz et al., 2010] introduces an automatic 3D geometry-based quantification of colocalization between two channels using three dimensional geometry structures of objects. The approach consists of two main steps. The first step is 3D spot detection step, where different 3D image filtering and smoothing operations are applied to obtain coarse spots and to reduce the noise. The image is convolved by a 3D-Gaussian filter with a standard deviation $\sigma_f$ proportional to the desired spot width. The second step is the spot quantification, where each of the previously detected spot candidate is evaluated and fitted to a 3D-Gaussian function using least-squares fitting model to specify the properties (e.g. size, structure, coordinates etc.). In image data the 3D intensity profile around a defined voxel (seed point specified using local maximum search) can be well represented by a 3D Gaussian function $g(x)$. Among different 3D Gaussian models $g_M$ with various parameters, the model with the lowest value for the objective function:

$$\sum_{x \in ROI} (g_M - g(x))^2$$

is chosen as the best fitted Gaussian function. The parameters of the best selected function specifies and defines the properties of the region. The computation time is relatively short (approximately 3s - 5s) with a good performance. Unfortunately this tool is not freely available.

Ram et al. [Ram et al., 2010] developed a method to segment and classify 3D spots in fluorescence images. They applied this approach on two application fields. The algorithm mainly consists of two steps spot segmentation and spot detection. The spot segmentation consists of 3D smoothing, top-hat filtering, intensity thresholding and 3D region growing. The spot detection is based on machine learning approaches. After spot segmentation a number of discriminative features (e.g. volume, contrast, texture etc.) are extracted from them and based on these features a classifier is generated to classify the spots as either true or false spots (spot detection phase).

Raimondo et al. [Raimondo et al., 2005] proposed a multistage algorithm for automated classification of FISH images from breast carcinoma samples. The algorithm consists of two stages: nucleus segmentation and spot segmentation using different techniques. The nucleus segmentation step consists of a nonlinear correction, global thresholding and marker-controlled watershed transformation. The spot segmentation step consists of top-hat filtering, followed by thresholding and gray-level template matching to separate real signals from noise.

Bolte et al. [Bolte and Cordelieres] introduced an object-based analysis to detect and segment spots in FISH images. Furthermore they provide an online available *IMAGEJ* tool called three-dimensional object counter17 that uses the object-based colocalization analysis and allow an automated colocalization analysis in a three- dimensional space. In the segmentation phase the foreground regions are emphasized using a 3D anisotropic smoothing to remove the effect of noise, 3D-top hat filter to represent the foreground details better, binary thresholding and 3D region growing. In the classification phase the resulting segmented spots are classified into two classes: true spots and false positives.

Netten et al. [Netten et al., 1996] used cell nucleus in slides of lymphocytes from a blood culture and introduced an automatic counting of spots. Their methods consists of three steps:
1) Filtering to suppress noise and applying a global threshold;
2) Detection of nuclei in using morphological filters;
3) Segmentation of hybrization spots within the nucleus using a nonlinear filter and top-hat transform.
Their performance in spot detection is acceptable, but often this method often yields segmented spot regions that contain mislabeled pixels near the borders of the spots.

Lerner et al. [Lerner B et al., 2007] proposed a FISH image classification system which is not limited on just dot-like signals and provides a methodology for segmentation and classification of dot and non-dot signals. This method is based on the properties of in-focus and out-of-focus images captured at different focal planes. They were used initially for the classification a neural network (NN)-based algorithm (well described in [Hastie et al., 2009]) and later provided a Bayesian classifier in- stead of NN, to avoid dependency on a large number of parameters and the NN architecture settings.

# 2 METHODS

We introduce the 3D-OSCOS (3D-**O**bject **S**egmentation and **CO**localization Analysis based on **S**patial Statistics) algorithm which is implemented as a user-friendly toolbox for interactive detection of 3D objects and visualization of labeled images.

The well-developed 3D-OSCOS toolbox, as depicted in Figure 1, consists of five main steps. Each individual step of this workflow may crucially affect the study result. This workflow starts with image acquisition and ends with a set of detected objects which is issued both visually and as a list of individual objects with their statistical properties. According to the proposed workflow in [Smal et al., 2010] where small subcellular objects were detected, the 3D-OSCOS workflow (Figure 1) is organized as follows:

1. Image acquisition includes all steps from capturing images to forming a digital image data set.

2. Image preprocessing refers to all types of manipulation of acquired image, resulting in an optimized output of the image.

3. Segmentation includes all steps to subdivide an image in some connected meaningful regions referred to as objects. In general this step requires a priori knowledge on the nature and content of the images, which must be integrated into algorithms.

4. Colocalization analysis refers to the investigation of the spatial overlap between two or more sub-cellular objects of different image channels.

5. Statistical analysis sums up all descriptive and spatial statistic methods to measure and describe the properties of the segmented sub-objects.

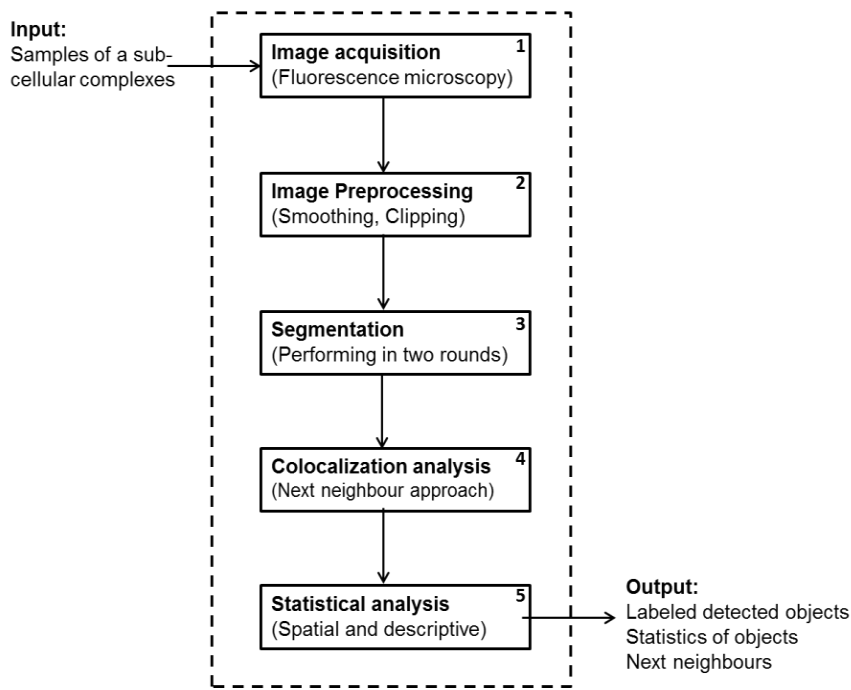

Figure 1: Workflow of 3D-OSCOS process. As it is illustrated, after sample preparation 3D-OSCOS performs 5 steps of image acquisition, image preprocessing, segmentation, colocalization and spatial analysis. The outputs are a set of labeled detected objects, their statistics and the next neighbor statistics.

*Image Acquisition*

As sketched in Figure 1, the process of colocalization analysis begins with the image acquisition. This step deals with the choice of the microscope and correct image acquisition settings. It should be bared in mind that some crucial instructions of image acquisition are essential for the whole pipeline. These points should be taken into account before the image is acquired by the microscope and after that post processing by computer. Understanding image acquisition techniques and their effects will help to design better experiments and improve the validity and reproducibility of quantitative image analysis.

For the image acquisition step commonly a charge-coupled device (CCD) camera is used to acquire and record emitted light signal. CCD is a small, centimeter-size chip of silicon that is divided up into millions of tiny picture elements which is able to store photo-electronic signals. The efficiency of light collection is so great that even weak images can be recorded in just a few milliseconds [Murphy, 2001].

There are various terms that define imaging performance. These criteria can be categorized as follows:
- Spatial resolution (ability to capture fine details without seeing pixels)

- Signal-to-noise (clarity and visibility of object signals in the image)

- Numerical Aperture (the intensity of the signal captured by the microscope)

- Dynamic range (number of resolvable steps of light intensity, gray-level steps)

- Sampling (Fitting a single sub-resolution light source to an appropriate number of pixels on the detector to avoid over- or under-sampling).

**Image Processing**

After visual inspection, it can be recognized that the desired 3D subcellular objects appear as bright spot-like peaks with a 3D-Gaussian form on a diffuse background light. Further due to limitations in imaging technology, a part of the spots are blurred and noisy. The brightness and contrast of the spots also varies and the illumination also differs over the slices. The main goal of the preprocessing phase is to reconstruct true fluorescence spots and to suppress the background noise as good as possible based on image processing approaches like filtering and clipping [Ronneberger et al., 2008].

Common image processing steps can be roughly divided into high-level image processing (e.g. image analysis or segmentation) which is more application-specific, and more general approach low-level image processing (e.g. histogram adjustment). Low-level processing denotes manual or automatic techniques, which can be realized without a-priori knowledge on a specific content of the image [Gonzalez and Woods, 2008].

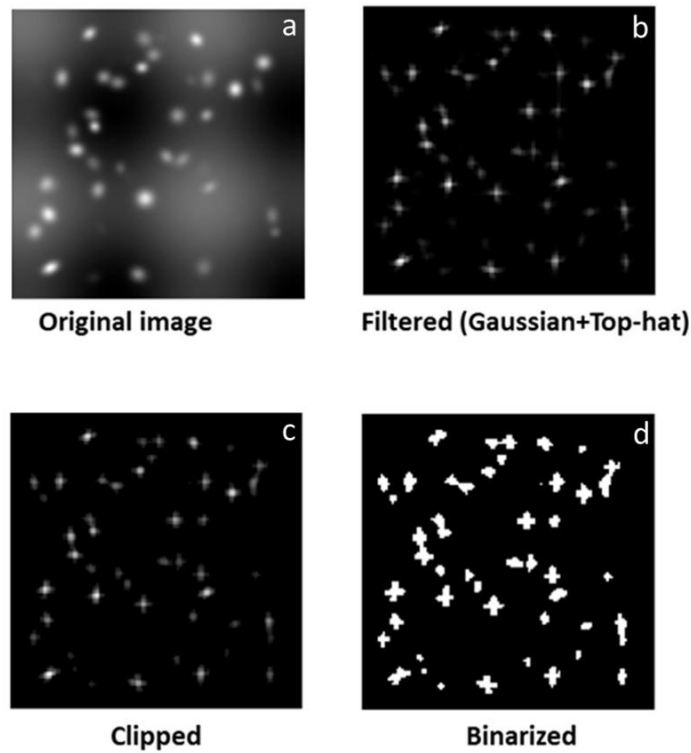

Figure 2: An illustration of image preprocessing results. A sample image layer (a) with the effect of filtering (b), clipping (c) and binarization (d).

Segmentation generally means dividing an image into connected regions corresponding to objects. The segmentation is usually based on identifying common properties.

Inspecting provided 3D Fluorescent images shows that two kinds of heterogeneities can be observed:

- Cell distribution heterogeneity: Various number of spots in different cell regions.

- Spot heterogeneity: Various spot volume size, spot average intensity and their pairwise distance).

We noted that both the inter-object distances and signal intensities differ in various regions of sub-cellular compartments. Therefore we suggest performing the segmentation phase in multiple rounds. First the objects are segmented and in a further step among all detected objects, those with a size greater than a defined threshold are analyzed separately in a further round to divide them into several smaller sub-objects fulfilling a user-defined range. Figure 3 shows a sketch that objects can be grouped into two types. The objects are either well separated or some of them are very closed-by and connected forming big-objects" which should be segmented in two separated segmentation routines, respectively.

**First round: (3D connected component labeling)**

To find all connected components, the topological relationship of adjacent pixels are analyzed. In other words, for each pixel the 26 voxels in the three dimensional neighborhood are inspected and all adjacent pixels above a certain threshold are considered to be a part of the same structure as the reference pixel. All pixels which are considered as a connected component are tagged or labeled with the same number.

To define an appropriate threshold, the well-known global thresholding [Otsu, 1979] is used which seeks to maximize between class variance [Gonzalez and Woods, 2008]. Due to the unequal intensity distributions among the slices, the global thresholding is applied for individual slice separately to get a threshold which is independent on the intensity distribution of other slices. The user has also the opportunity to affect the threshold by adding a factor which will be multiplied to the global threshold (called threshold factor c2).

Thus for each slice i, first the global threshold is determined and then multiplied by the factor c2 which is defined by the user in advance:

$$T_{binarization} = \text{globalThreshold}(\text{slice}(i)) * C_2$$

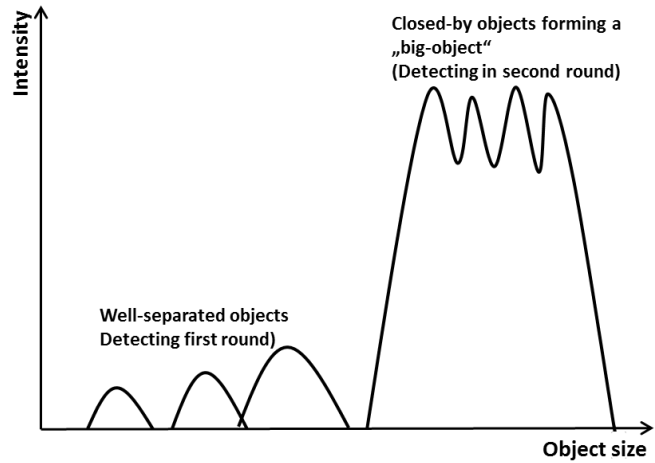

Figure 3: A sketch of well separated and close-by objects (forming a big-objects") based on the intensity and object size.

An appropriate threshold for each individual slice is determined by using a global thresholding approach and by considering the user defined a thresholding factor. All voxels with an intensity above the threshold will be considered as potential object's pixel other pixels will be considered as being part of the background. After binarization, all slices will be scanned from top left corner of first slice to the lower right corner of the last slice in order to find 3D connected components. Each time a new object voxel with an intensity greater than zero is found, its 26 neighbors (9 neighbors on the upper and lower slice respectively plus 8 neighbors on the same slice) are checked for a voxel which doesn't belong to the background that can be interpreted as its connected component. Each defined object (a set of connected components) is tagged by a unique number.

The following pseudo code outlines the first round of segmentation task implemented in MATLAB function *"lableObjects.m"* as follows:

"

> **First round of segmentation:**
>
>   **Input:** Threshold factor $c_2$, ObjSize=expected object size
>
>   1. For each slice **i**, calculate the binarization threshold $T_{binarization}$ as above.
>   2. For each pixel on each slice, check if $\mathbf{f}(x, y, z) \geq T_{\text{binarization}}$, then map it to 1, otherwise to zero.
>   3. For each nonzero voxel, check the 26-neighborhood for connected components and define all voxels which are connected in 26 neighborhood as connected components.
>   4. Label each individual found connected component (refer to object) with an unique number
>   5. Determine for each detected object its properties, e.g. size, coordinates etc.
>   6. Divide all detected objects into two categories of objects:
>      Normal objects: All objects with a size between ObjSize and 3×ObjSize
>      Big objects: All objects with a size of greater than 3×ObjSize
>
>   **Output:** A list of all detected objects with their individual statistical properties.

**Second round: (Local maximum search with distance)**

The first round of segmentation depends on the volume size of the detected objects, following two categories of objects which can be defined:
  • Normal size Object
  • Big Size Object

where first category consists of objects which have a size in user-defined range. All objects of this group are added to the end list of detected objects. The big object category includes objects which are many time greater than the user defined expected object size (most of them are located normally in the X chromosomal space). The main task is to resolve the big objects in biological meaningful small sub-objects satisfying the defined object size range, assuming that each of them is derived from a single light emitting molecule.

The second round of segmentation deals with the task of dividing big objects into sub-objects, for this task a novel approach which is an extension of the well-known local maximum search by taking into account their inter-distance based on Euclidean distance, is introduced. First for each big object the number of desired sub-objects (N) should be determined depending on its size and the size of the expected objects. After that all 3D local maximum points are searched, i.e. those points which have the highest intensity within a local 3D window (26 neighbors). Then the algorithm looks for n brightest 3D local maximum points satisfying a minimum distance criterion. At the end of this step, each detected voxel fulfilling both conditions are acted as seed points and represent reference locations for the desired objects. Based on the found seed points, all remaining voxels should be assigned to one of these points to form a sub-object. In order to specify each voxel which point is the most suitable seed point, the next neighbor approach is used and hence each remaining voxel is assigned to that seed point with the smallest Euclidean distance.

The following pseudo code outlines the second round of segmentation which performs dividing of big-objects in smaller sub-objects and is implemented in "*analyseHuge-Objects.m*" and "*findDistancedMaxima.m*" as follows:

---

**Second of segmentation:**
**Input:** List of big objects, ObjSize = expected object size, d = suitable object distance

1. For each big object (i) specify the desired number of sub-objects $N(i)$ as a function of big object size and user defined expected object size:
$$N(i) = \text{big object size}(i) / \text{ObjSize}.$$

2. Determine all 3D local maximum points $P_1 \ldots P_n$ of object (i).

3. Find and identify the brightest point ($P_1$) of the big object.

4. Search for the next brightest point ($P_2$).

5. Check whether the Euclidean distance between $P_1$ and $P_2$ is greater than $d$ or an Euclidean distance is greater than $d/2$ and at the same time it is on the same slice or not.

6. If the condition is fulfilled, add $P_2$ to the list of seed points, otherwise look for the next brightest 3D local maximum points $P_3 \ldots P_n$ which fulfil the conditions.

7. Repeat steps till a number of $N$ seed points are found.

8. For each remaining voxel, check based on the Nearest neighbor approach which seed point is the nearest point.

9. Assign each voxel to its nearest seed point to form individual sub-objects.

10. Label each individual found connected component (refer to object) with an unique number.

11. Determine for each detected object its properties, e.g. size, coordinates etc.

**Output:** A list of all detected sub-objects with their individual statistical properties

---

# 3 RESULTS

In order to quantify the performance of the 3D-OSCOS algorithm and to compare it against a ground truth reference, a set of simulated (computer-generated) images which consists of 3D images with different contrasts and signal-to-noise ratios was used. Further the introduced algorithm was compared to other algorithms proposed by Bolte et al. [Bolte and Cordelieres], e.g. the ImageJ plugin Object Counter3D in terms of accuracy and processing time. Moreover the 3D-OSCOS method was validated by applying an experimental data set and compared visually its output to the manual quantification result. For the performance measurement, the commonly used metrics proposed in [Fawcett, 2006] were applied as follow: First, a true positive (TP) is defined as a correctly founded object, and a false positive (FP) is a detected object for which there is no match in the reference image. A false negative (FN) corresponds to a missing object in the detection result. The same definitions may also be applied for pixel-level analysis described in [Ruusuvuori et al., 2010].

In computer generated images the number and location of the spots are known, as they are simulated by computer with user-defined properties. Recently, benchmark image collections of cellular biological samples have been developed in order to facilitate comparison and validation of object detection and image analysis methods: [Ljosa et al., 2012], [Ruusuvuori et al., 2010] and [Drelie Gelasca et al., 2009]23. Un- fortunately, the images are provided only as two-dimensional images, thus they are not suitable for our investigation. Therefore for the evaluation of our toolbox, we used a simulated 3D image developed by Wörz et al. [Worz et al., 2010].

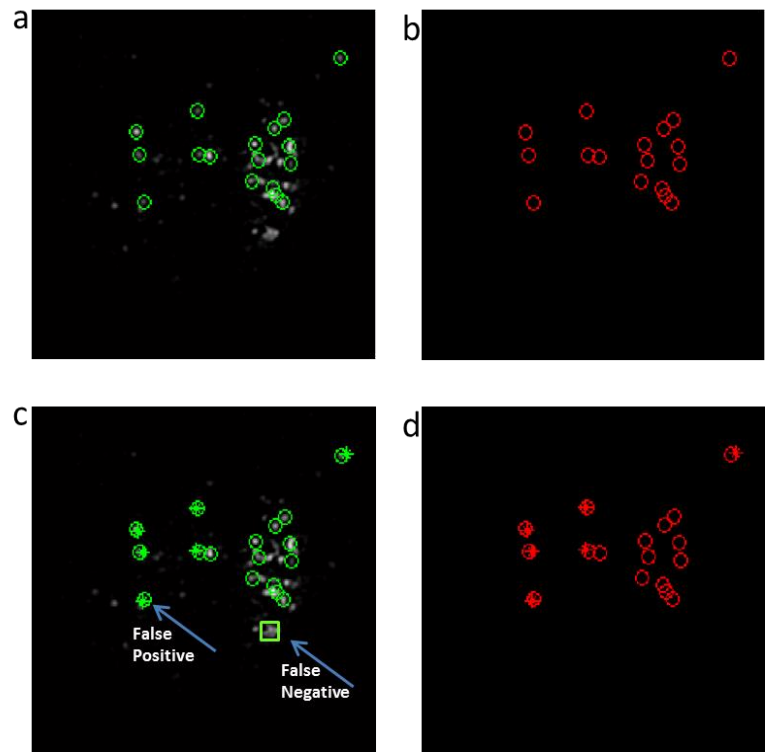

Figure 4: Visual output of labeled objects. Showing on a single layer of the 3D MSL2 image data set, where a) and b) show all automatically detected objects represented by a green circle on an original and black background respectively. c) and d) illustrates a combination of labels including manual post-labeling (False Negative) and manual deleting (False Positives).

The provided three-dimensional artificial data set consists of 16 slices with a width and height of 128 pixels, respectively. The dynamic range of each voxel is 16 bit and the intensity of the voxels are very close together range between 9,960 to 10,040.

The three-dimensional artificial image consists of exactly one hundred objects of a certain range of size distributed randomly over an 3D image. The image is provided in six different versions relating to the signal-to-noise relation. As depicted in Figure 5 the original image without any noise (Fig. 5) and further images which are generated by adding different levels of random noise to the original image (Fig. 5). Thus all images consists of objects which are equally dispersed over them but with various signal-to-noise ratios.

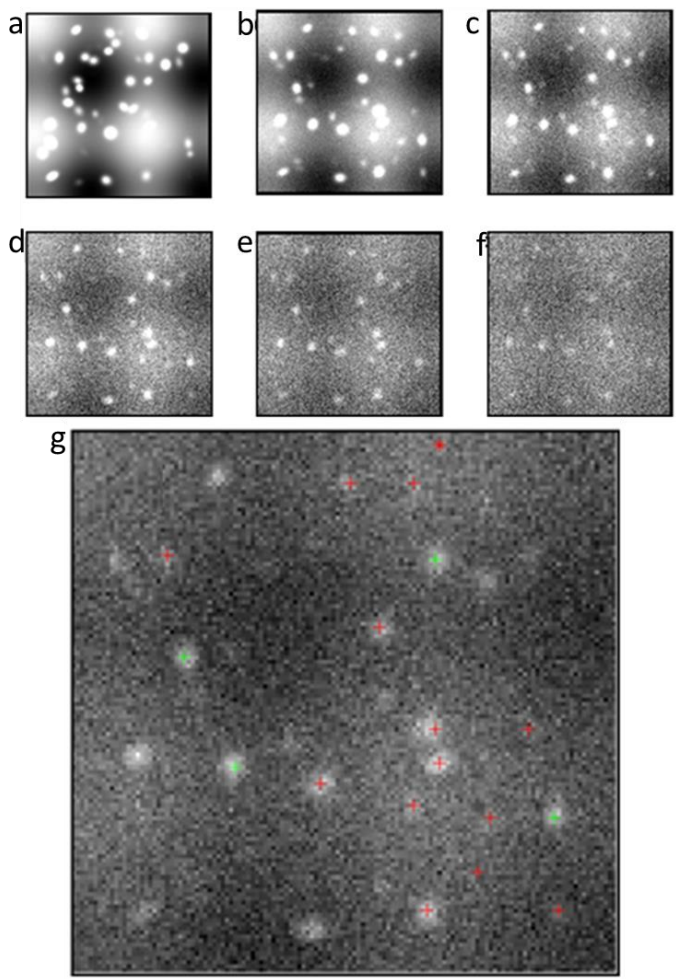

Figure 5: A sample slice of artificial image data. (a) Artificial image without noise, (b-f) Stepwise increasing of noise on the same image data. (g) The output of labeling process using 3D-OSCOS on the same slice.

In order to measure and quantify the performance of the proposed method, a clear defintion of ground truth is required. Thus in case of real data set, we define the manual detected objects as ground truth and in case of artificial data set, the number of objects is known since it is generated based on user inputs. Therefore we are able to compare our algorithm to other 3D object detection methods based on evaluation of the detection performance against a known number of ground truth.

First we analyzed the performance based on artificial data set. As it is shown in Table 8.4 the performance of 3D-OSCOS depends on the signal-to-noise ratio of each image type. The higher the noise the lower the performance where all for all types of images the execution time is nearly the same. It should be mentioned that during the experiments some objects were indicated as double-labeled, it means that some object that are expanded over several slices have become more than one mark. Therefore the final task was to remove all double-labeled by checking for unique number of marks in each object. It was performed based on Euclidean distance between the marks for each unique object.

| Data set | Number of labeled objects | True Positive rate | False Positive rate | False Negative rate | Calculation time |
|---|---|---|---|---|---|
| **N00** | 109 | 99% | 9% | 1% | 3s |
| **N03** | 109 | 97% | 9% | 3% | 3s |
| **N10** | 112 | 97% | 15% | 3% | 3s |
| **N20** | 104 | 96% | 5% | 4% | 3s |
| **N30** | 95 | 91% | 7% | 9% | 3s |
| **N50** | 117 | 80% | 15% | 20% | 5s |

Table 1: The performance of 3D-OSCOS using artificial image. The number of detected object for each type of artificial image is listed. Based on these values the performance can be determined by giving the TP-, FP- and FN-rates. Each computer-generated image consists of exact 100 objects. For example for the first data set (N00), our tool detected 109 objects, 99 of those are correctly detected and one object is not detected. Therefore, the TP-rate equals $99/100 = 99\%$, the FN-rate is $1/100 = 1\%$ and finally the FP-rate is equal to $10/109 = 9\%$. In addition the calculation or the running time of the automatic object detection is given.

The proposed algorithm is evaluated with respect to the spot detection task. We compared the 3D-OSCOS to two other algorithms proposed by Bolte el al. [Bolte and Cordelieres]. The number of correctly detected objects using 3D-OSCOS is much closer to the real number compared to the Object counter plugIn (see Figure 6).

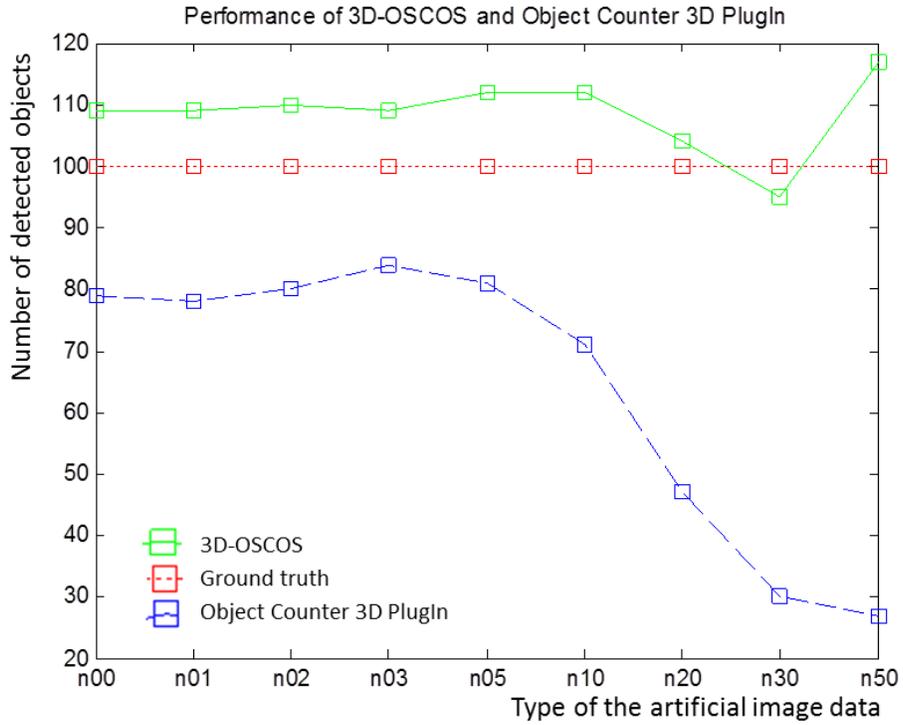

Figure 6: Performance of two different methods using artificial image data. The red line shows the real number of objects in the image data. The green line related to the 3D-OSCOS is very close to the real number and in contrast the performance of the object counter plugin decreases by increasing the noise ratio of the image data.

## *CONCLUSION*

Due to the huge amount of data being generated by the current generation of microscopes and the need of accurate statistical investigations, biologist lacked automated methods for quantitative analysis of cell nucleis and in particular for detecting of sub-cellular objects in microscope images. For these purposes, some problem-specific methods have been developed, where most of them are applicable just for 2D image data like Lerner et al. [Lerner B et al., 2007] and Raimondo et al.[Raimondo et al., 2005] (for an overview and evaluation review see [Ruusuvuori et al., 2010]).

We have developed an fully automatically 3D object detection toolbox which provides a user-friendly interface in order to interact with the user for parameter setting inputs and to prevent a "black box" effect. The 3D-OSCOS detects automatically objects in three-dimensional images with an additional option for the user to check the result visually and if necessary to enhance it. An advantage feature of 3D-OSCOS is the option to run the program either fully automatic when the user has no prior information about the image or semi-automatic when the user wants to contribute his knowledge about the image data into the detection process. The statistical power and utility of this method is shown by comparing the detection power in comparison to other freely available methods.

We established an extension of object detection and colocalization analysis in the sense that the statistical spatial analysis is used in order to analyze the full 3D spatial information about objects, their localizations and interactions. To achieve this goal, after detecting 3D objects and determining their statistical properties, each of them is presented by its center point in order to analyse their potential interactions based on well-developed statistical spatial approaches.

Another notable finding of this work, according to the user intervention option during the detection phase, is that the inaccurate and time consuming task of manual object detection can be avoided as follows: The main part of the 3D detection phase can be executed automatically (e.g. 95% - 98% of the total workload) and for the last detailed optimizations the user has the opportunity to inspect visually the result and enhance it when needed.

All findings and experiences in this field are freely available for the community as an open source MATLAB and R packages and further detailed documentations